\documentclass[10pt,conference]{IEEEtran}
\IEEEoverridecommandlockouts
\usepackage{textcomp}
\usepackage{verbatim}
\usepackage{color}
\usepackage{xcolor, colortbl}
\definecolor{graytwo}{gray}{.7}
\usepackage{xspace}
\usepackage{balance}
\usepackage{tabulary}
\usepackage[T1]{fontenc}
\usepackage{array}
\usepackage{diagbox}

\usepackage{array} 
\usepackage{multirow}
\usepackage{booktabs}
\usepackage{tabularx}
\usepackage{caption}
\usepackage[ruled,lined,linesnumbered,vlined,algo2e]{algorithm2e}
\usepackage{amsmath}
\usepackage{subcaption}
\usepackage{listings}
\usepackage{lipsum}
\usepackage{hyperref}
\usepackage{cleveref}
\usepackage{wrapfig}
\usepackage{graphicx}
\usepackage[T1]{fontenc}
\usepackage[ansinew]{inputenc}
\usepackage[many]{tcolorbox}
\usepackage{extarrows}
\usepackage{booktabs}
\usepackage{tabularx}
\usepackage{enumitem}
\setitemize[0]{leftmargin=15pt}
\usepackage[justification=centering]{caption}
\usepackage{pifont}
\usepackage{makecell,multirow,diagbox,rotating}
\usepackage{longtable}
\usepackage{etoolbox} 
\usepackage{algorithm}
\usepackage{algorithmicx} 
\usepackage{algpseudocode}
\usepackage{amsmath}
\usepackage{amsfonts}
\usepackage{amssymb}
\usepackage{dutchcal}
\usepackage{breakurl}
\definecolor{codegreen}{rgb}{0,0.6,0}
\definecolor{codegray}{rgb}{0.5,0.5,0.5}
\definecolor{codepurple}{rgb}{0.58,0,0.82}
\definecolor{backcolour}{rgb}{0.95,0.95,0.92}

\lstdefinelanguage{JavaScript}{
  keywords={typeof, new, true, false, catch, function, return, null, catch, switch, var, if, in, while, do, else, case, break, const, prototype},
  keywordstyle=\color{violet}\bfseries,
  ndkeywords={class, export, boolean, throw, implements, import, this},
  ndkeywordstyle=\color{darkgray}\bfseries,
  identifierstyle=\color{black},
  sensitive=false,
  comment=[l]{//},
  morecomment=[s]{/*}{*/},
  commentstyle=\color{gray}\ttfamily,
  stringstyle=\color{blue}\ttfamily,
  morestring=[b]',
  morestring=[b]",
  frame=none,
  numbers=none
}

\hypersetup{
  colorlinks   = true,    
  urlcolor     = blue,    
  linkcolor    = blue,    
  citecolor    = blue      
}

\usepackage{fancybox}


\makeatletter
\renewcommand{\maketag@@@}[1]{\hbox{\m@th\normalsize\normalfont#1}}%
\makeatother

\usepackage{makecell}
\usepackage{threeparttable}
\makeatletter  
\newif\if@restonecol  
\renewcommand\footnoterule{%
	\kern-3\p@
	\hrule\@width\columnwidth
	\kern2.6\p@}

\usepackage{url}

\SetCommentSty{mycommfont}

\definecolor{Green}{RGB}{0,180,0}
\usepackage{xspace} 

\usepackage[font=small,skip=2pt]{caption}
\newcommand{\distance}{2pt}
\definecolor{darkgrey}{HTML}{434343}
\newtcolorbox{mybox}[2][]{text width=0.95\linewidth,fontupper=\normalsize,
fonttitle=\bfseries\sffamily\scriptsize, colbacktitle=darkgrey,enhanced,
attach boxed title to top left={yshift=-2mm,xshift=3mm},
boxed title style={sharp corners},top=4pt,bottom=2pt,left=2pt,right=2pt,
  title=#2,colback=white}

\begin{document}

\title{AI-Driven Strategies for Reducing Student Withdrawal \\ \LARGE A Study of EMU Student Stopout}


\author{\IEEEauthorblockN{Yan Zhao\IEEEauthorrefmark{1}, 
Amy Otteson \IEEEauthorrefmark{1}}
\IEEEauthorblockA{\IEEEauthorrefmark{1}\textit{Eastern Michigan University}, USA}
}

\maketitle
\thispagestyle{plain}
\pagestyle{plain}


\begin{abstract}
Not everyone who enrolls in college will leave with a certificate or degree, but the number of people who drop out or take a break is much higher than experts previously believed. In December 2013, there were 29 million people with some college education but no degree. That number jumped to 36 million by December of 2018, according to a new report from the National Student Clearinghouse Research Center\cite{NSCR}.

It is imperative to understand the underlying factors contributing to student withdrawal and to assist decision-makers to identify effective strategies to prevent it. By analyzing the characteristics and educational pathways of the stopout student population, our aim is to provide actionable insights that can benefit institutions facing similar challenges. 

Eastern Michigan University (EMU) faces significant challenges in student retention, with approximately 55\% of its undergraduate students not completing their degrees within six years. As an institution committed to student success, EMU conducted a comprehensive study of student withdrawals to understand the influencing factors. And the paper revealed a high correlation between certain factors and withdrawals, even in the early stages of university attendance. Based on these findings, we developed a predictive model that employs artificial intelligence techniques to assess the potential risk that students abandon their studies. These models enable universities to implement early intervention strategies, support at-risk students, and improve overall higher education success.

\end{abstract}
\author{\IEEEauthorblockN{Yan Zhao\IEEEauthorrefmark{1}, 
Amy Otteson \IEEEauthorrefmark{1}}
\IEEEauthorblockA{\IEEEauthorrefmark{1}\textit{Eastern Michigan University}, USA}
}

\maketitle

\section{Introduction}

Student withdrawal from college is a critical issue that significantly impacts higher education institutions across the United States. Despite growing enrollment rates, a considerable number of students leave college without completing their degrees. This trend has raised concerns among educational policy makers and institutions. 

Addressing the issue of student withdrawal is essential for improving graduation rates and ensuring student success. By identifying the underlying factors that contribute to student withdrawal and developing strategies to address these factors, institutions can better support their students. This study aims to provide a comprehensive analysis of the stop-out student population at Eastern Michigan University (EMU) to understand the characteristics and educational pathways that lead to withdrawal.

Specifically, this study focuses on two primary research questions concerning new First-Time In Any College (FTIAC) students. 
\textbf{
\begin{itemize}
    \item Q1: Which students are most likely to withdraw from school, and what common characteristics do they share?
    \item Q2: Can these common characteristics be used to predict the risk of student withdrawal?
\end{itemize}
}

For Q1, the analysis includes various factors such as academic performance, demographics, admission criteria, financial status, and other relevant characteristics to understand the predictors of student withdrawal.
In order to quantify the stopout group, retention rates are used as metrics. Retention rates measure the percentage of first-time undergraduate students who return to school in subsequent years. 
For Q2, the study examines whether the common characteristics identified in Q1 can be utilized to predict the risk of student withdrawal. By leveraging machine learning techniques, predictive models are developed to identify students at high risk of stopping out. These models use identified predictors, such as academic performance, demographics, institutional factors, engagement levels, and financial aid status, to assess the likelihood of student withdrawal. The goal is to provide actionable insights that enable EMU to implement targeted interventions and support strategies to improve student retention and success.

The data drawn from EMU's enrollment and degree records between Fall 2013 and Fall 2017 provide a foundation for exploring these questions. The insights gained from this analysis are expected to inform strategies that can improve student retention rates, not only at EMU but also at other institutions facing similar challenges.

\section{Related Work}

The issue of college student withdrawal has significant implications for both students and educational institutions. Extensive research has been conducted to understand the factors that contribute to student attrition and identify effective strategies to improve retention rates. This section reviews the relevant literature and previous studies on various dimensions of student withdrawal. 

Previous research consistently shows that demographic characteristics and academic performance are critical predictors of student withdrawal. Tinto \cite{tinto1993leaving} emphasized the importance of academic integration, suggesting that students who perform well academically and feel integrated into the academic community are less likely to withdraw. Herzog \cite{herzog2005measuring} found that GPA and credit hours completed are significant predictors of student persistence. 

Astin's theory of student involvement \cite{astin1984student} and Kuh et al.'s emphasis on high-impact educational practices \cite{kuh2008high} highlight the importance of student engagement in retention. 
Astin's theory of student involvement \cite{astin1984student} and Kuh et al.'s emphasis on high-impact educational practices \cite{kuh2008high} highlight the importance of student engagement in retention. 

Our study extends this by integrating correlation analysis to offer a holistic understanding of student withdrawal, combining quantitative rigor with qualitative depth. And instead of focusing on one single factor, our research utilizes a comprehensive dataset from Eastern Michigan University (EMU) to explore the interplay between academic performance, demographic characteristics, and admission criteria. 

Stratton, O'Toole, and Wetzel \cite{stratton2007dropout} found that students are most vulnerable to withdrawal during their first year of college. Pascarella and Terenzini \cite{pascarella2005how} noted that the first year is essential to establish academic habits and social connections.

Unlike previous studies, our research incorporates a predictive model to analyze the risk of student withdrawal at the early stage of university attendance, providing valuable information on potential at-risk students. 

These comprehensive approaches offer specific insights that can directly inform policy and practice at EMU and have broader applications for similar institutions.

Zhao \cite{zhao2024practice} utilized Markov Chain models for enrollment prediction, demonstrating the potential of advanced analytical techniques in educational research. While this research is useful for identifying early warning trends of withdrawal, it does not provide the capability to pinpoint individual at-risk students.

\section{Data Collection}

This study aims to analyze the factors that contribute to the withdrawal of students from Eastern Michigan University (EMU) and to identify common characteristics among students who are more likely to leave without completing their degree. 

We use retention rates as metrics to determine whether students stay or leave the school, rather than relying on degree completion rates. Retention rates measure the percentage of first-time undergraduate students who return to school in subsequent years. Avg FA \#2, Avg FA \#3, Avg FA \#4, Avg FA \#5, and Avg FA \#6 represent the average proportions of freshmen entering in Fall 2013 through Fall 2017 at EMU who returned in the second, third, fourth, fifth and sixth fall, respectively.
We chose not to use final degree completion rates directly because our goal is to identify at-risk students as early as possible. By doing so, the university can implement timely interventions to support these students and improve their chances of success.

Meanwhile,we collect data regarding academic performance, specifically term GPA and term earned Student Credit Hours (SCH), admission criteria like ACT score and decision GPA and demographic information. 

Data were derived from EMU's enrollment and degree records between Fall 2013 and Fall 2017. 

\begin{itemize}
    \item \textbf{Retention Rates}: Percentage of first-time undergraduate students returning in subsequent years.
    \item \textbf{Term GPA}: Average GPA of students categorized into five ranges: >3.5, 3.5-3.0, 3.0-2.0, 2.0-1.0, and <1.0.
    \item \textbf{Term Earned SCH}: Number of credit hours completed in the first fall term, categorized into four ranges: >12, 9-12, 6-9, and <3.
    \item \textbf{ACT Scores}: Composite scores categorized into four ranges: >30, 24-29, 18-23, and 12-17.
    \item \textbf{Decision GPA}: High school GPA used for college admissions.
    \item \textbf{Avg Fall \#2} : the average proportions of freshmen entering in Fall 2013 through Fall 2017 at EMU who returned in the second year.
\end{itemize}

\section{Correlaiton Experiments}
This section presents our investigation for the first research
questions RQ1. we will first introduce the
study method and then describe the results of our experiment.

\textbf{RQ1: Which students are most likely to withdraw from school, and what common characteristics do they share?} 

\subsection{Pearson Correlation Coefficient Calculation}
The Pearson correlation coefficient, denoted as \( r \), quantifies the linear relationship between variables:

\[ r = \frac{\sum (X_i - \bar{X})(Y_i - \bar{Y})}{\sqrt{\sum (X_i - \bar{X})^2 \sum (Y_i - \bar{Y})^2}} \]

Calculating the Pearson correlation coefficient provides several key benefits:

\begin{itemize}
    \item \textbf{Quantifies Relationships}: It gives a precise measure of the strength and direction of the relationship between term GPA and retention rates.
    \item \textbf{Statistical Validation}: Provides formal statistical validation of observed trends.
    \item \textbf{Standardized Comparison}: Allows for consistent comparison across different studies and datasets.
    \item \textbf{Confidence Intervals}: Offers insights into the reliability and precision of the correlation estimate.
    \item \textbf{Objective Assessment}: Minimizes bias by providing an objective, mathematical assessment of the relationship.
\end{itemize}

\subsection{Academic Standing (GPA and SCH)}

Academic standing is a significant predictor of whether students are likely to stop out. We measured academic standing using two categories: term GPA and term-earned SCH in the first fall. 

\begin{figure}[ht]
    \centering
    \includegraphics[width=0.5\textwidth]{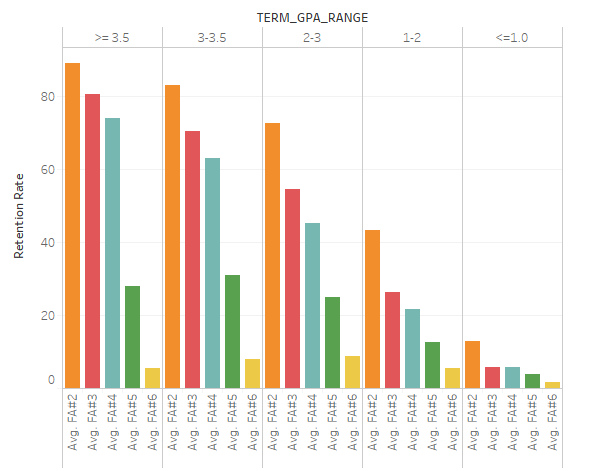}
    \caption{Retention Rates over Six Years by Term GPA}
    \label{fig:retention_rates_gpa}
\end{figure}

\subsubsection{Term GPA and Retention Rate Data}
Figure 1 shows the retention rates in different term GPA ranges over the first four years:

\begin{itemize}
    \item 89\%, 80\%, and 74\% for students with a term GPA > 3.5
    \item 83\%, 70\%, and 63\% for students with a term GPA between 3.5 and 3.0
    \item 73\%, 54\%, and 45\% for students with a term GPA between 3.0 and 2.0
    \item 43\%, 26\%, and 22\% for students with a term GPA between 2.0 and 1.0
    \item 13\%, 6\%, and 6\% for students with a term GPA below 1.0
\end{itemize}

Figure 2 shows the retention rates in the different term-earned SCH groups over the next four years are as follows:
\begin{itemize}

\item 76\%, 65\%, and 58\% for students who completed more than 12 credit hours in the first fall
\item 65\%, 49\%, and 42\% for students with term-earned SCH between 9 and 12
\item 41\%, 30\%, and 26\% for students with term-earned SCH between 6 and 9
\item 12\%, 8\%, and 9\% for students with term-earned SCH below 3
\end{itemize}
\begin{figure}[ht]
    \centering
    \includegraphics[width=0.5\textwidth]{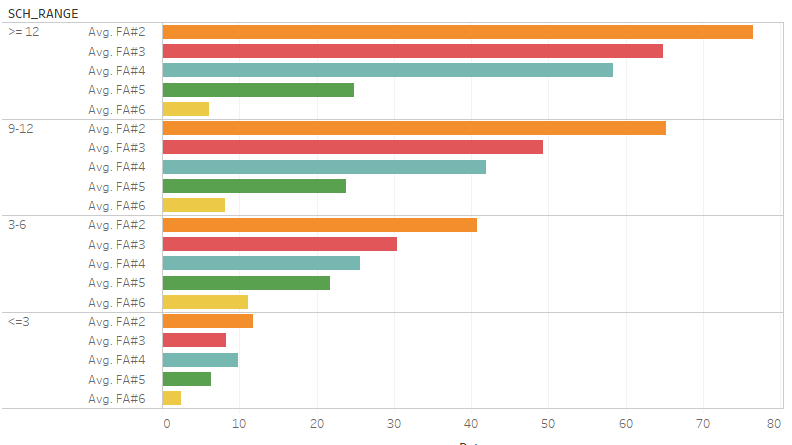}
    \caption{Retention Rates over Six Years by SCH}
    \label{fig:retention_rates_sch}
\end{figure}

\subsubsection{Correlation Results}
To calculate the correlation, we use the following data points:

\begin{itemize}
    \item For Term GPA and Retention Rates:
    \begin{itemize}
        \item Term GPA: [4.0, 3.25, 2.5, 1.5, 0.5] (representing GPA ranges: >3.5, 3.5-3.0, 3.0-2.0, 2.0-1.0, <1.0)
        \item Second FAll Retention Rates: [0.89, 0.83, 0.73, 0.43, 0.13]
    \end{itemize}
    \item For Term Earned SCH and Retention Rates:
    \begin{itemize}
        \item Term Earned SCH: [13.5, 10.5, 7.5, 1.5] (representing SCH ranges: >12, 9-12, 6-9, <3)
        \item Sencond Fall Retention Rates: [0.76, 0.65, 0.41, 0.12]
    \end{itemize}
\end{itemize}

Using these data points, we calculate the Pearson correlation coefficients.

\begin{tcolorbox}[colframe=black, colback=white, sharp corners=all, boxrule=1pt]
\subsubsection*{Correlation between Term GPA and Retention Rates}
\textbf{Pearson correlation coefficient:} \( r = 0.993 \) \\
\textbf{95\% confidence interval:} \( (0.979, 1.008) \)
\end{tcolorbox}

\begin{tcolorbox}[colframe=black, colback=white, sharp corners=all, boxrule=1pt]
\subsubsection*{Correlation between Term Earned SCH and Retention Rates}
\textbf{Pearson correlation coefficient:} \( r = 0.993 \) \\
\textbf{95\% confidence interval:} \( (0.974, 1.012) \)
\end{tcolorbox}

These high correlation values indicate a very strong positive relationship between both term GPA and retention rates, and term earned SCH and retention rates. This suggests that higher term GPAs and higher term earned SCH are strongly associated with higher retention rates at EMU. The rates at which students returned to EMU increased with the quality and amount of academic progress made during the first fall.

\subsubsection{Key Insights}
Several key insights can be gained regarding the relationship between academic standing and student retention rates over the first four years:

\begin{itemize}
    \item \textbf{Higher GPA and SCH Correlates with Higher Retention Rates}: Students with higher term GPAs and SCH in their first fall semester are significantly more likely to stay enrolled in subsequent years.
    \item \textbf{Declining Retention with Lower GPA and SCH}: There is a clear stepwise decline in retention rates as the term GPA and SCH decrease.
    \item \textbf{Severe Drop in Retention for Lowest GPA Bracket}: Students with term GPAs between 2.0 and 1.0 have markedly lower retention rates, with only 43\% returning in the second year. The situation is most critical for students with a term GPA below 1.0, where retention rates are dramatically low at 13\% in the second year. Students with SCH below 6 show the same trend.
\end{itemize}

These insights suggest that academic performance in the first term is a strong predictor of student retention. Targeted interventions could be especially beneficial for students with GPAs below 2.0 or SCH below 6 to help improve their academic standing and increase their likelihood of persisting in their studies.

Understanding these patterns can help universities allocate resources more effectively. Institutions can prioritize academic support services for students identified as high-risk based on their first-term GPA and SCH.

In summary, the data in Figures 1 and 2 highlight the critical importance of early academic success in predicting student retention, underscoring the need for early intervention strategies to support at-risk students and improve overall retention rates.

The difference in retention rates in the fifth and sixth falls is not as significant as in previous years because students with higher term GPAs and more term earned SCH have higher completion rates within four years.

\subsection{Admission Criteria (Test Scores and Decision GPA)}

Test scores and decision GPA (high school GPA) are two important college-admissions criteria.
\begin{figure}[ht]
    \centering
    \includegraphics[width=\linewidth]{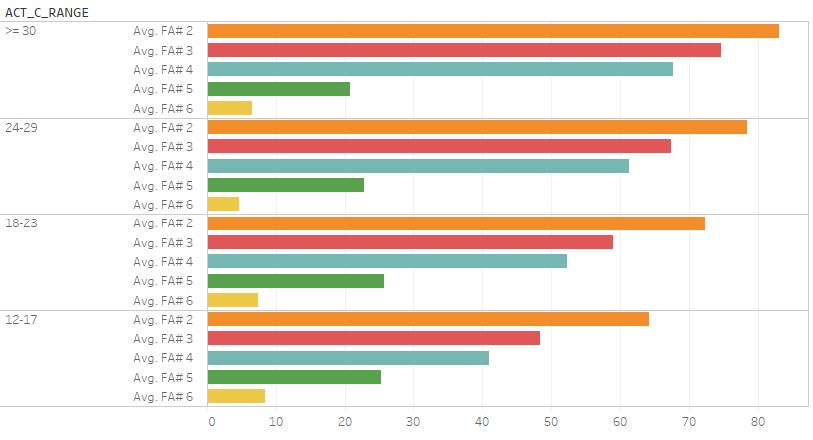}
    \caption{Retention Rates over Six Years of ACT Score Group}
    \label{fig:retention_rates_gpa}
\end{figure}

As shown in Figure 3. Specifically, 83\% students with an ACT Composite score greater than 30 re-enrolled in the second fall, while 78\% students with ACT Composite score between 24 and 29 were retained. Of the students whose composite ACT score was between 18 and 23, 73\% of the students reenrolled. Conversely, only 64\% students with ACT Composite score between 12 and 17 re-enrolled. They also keep the same trend in the next four fall terms. The retention rates for the third fall were 75\%, 67\%, 59\%, and 48\% and 68\%, 61\%, 52\%, and 49\% for the fourth fall for the four groups of undergraduate students, respectively.

\begin{tcolorbox}[colframe=black, colback=white, sharp corners=all, boxrule=1pt]
\subsubsection*{Correlation between ACT score and Retention Rates}
\textbf{Pearson correlation coefficient:} \( r = 0.992 \) \\
\textbf{P-value:} p=0.0075
\end{tcolorbox}

The correlation coefficient r = 0.992 indicates a very strong positive correlation between the ACT composite scores and retention rates for the second fall term. The p-value p = 0.0075 is well below the common significance threshold of 0.05, indicating that this correlation is statistically significant. 

\begin{figure}[ht]
    \centering
    \includegraphics[width=0.5\textwidth]{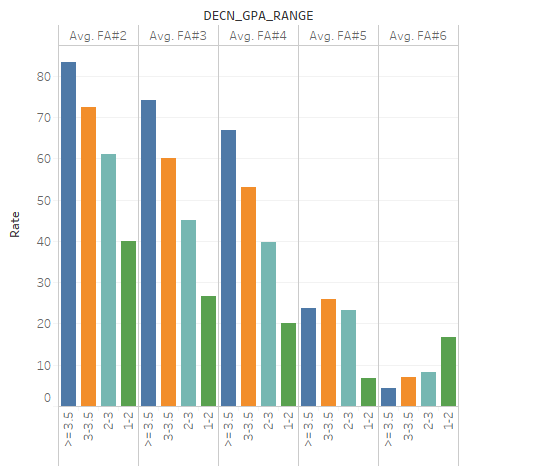}
    \caption{Retention Rates over Six Years by Decision GPA}
    \label{fig:retention_rates_sch}
\end{figure}

Figure 4 below shows the distribution of retention rates in different decision GPA groups. Specifically, we see that the retention rates in the first four years are 83\%, 74\%, and 67\% for students with a decision GPA greater than 3.5, decreasing to 72\%, 60\%, and 53\% for students with decision GPA between 3.0 and 3.5. For students with a decision GPA between 2.0 and 3.0, 61\%, 45\%, and 40\% were retained. There was a large drop in retention rates for students with a decision GPA of 1.0 to 2.0: 40\%, 27\% and 20\%. From the fifth fall, the retention rates are not consistent with previous years because students with higher ACT Composite scores or better decision GPA are more likely to earn their degree in four years.

\begin{tcolorbox}[colframe=black, colback=white, sharp corners=all, boxrule=1pt]
\subsubsection*{Correlation between Decision GPA and Retention Rates}
\textbf{Pearson correlation coefficient:} \( r = 0.995 \) \\
\textbf{P-value:} p=0.0049
\end{tcolorbox}

This very high Pearson correlation coefficient 0.995 and low p-value (< 0.05) indicate that the correlation between decision GPA and retention rates in the first year is statistically significant, strongly supporting the hypothesis that higher decision GPAs are associated with higher retention rates. 

Based on these insights, institutions can design targeted interventions to support students with lower ACT scores, potentially improving their retention rates. Admission policies can take these findings into account to ensure that students who may need additional support are identified early and provided with the necessary resources to succeed and stay enrolled.

\subsection{Demographic Information}
\begin{figure}[ht]
    \centering
    \includegraphics[width=\linewidth]{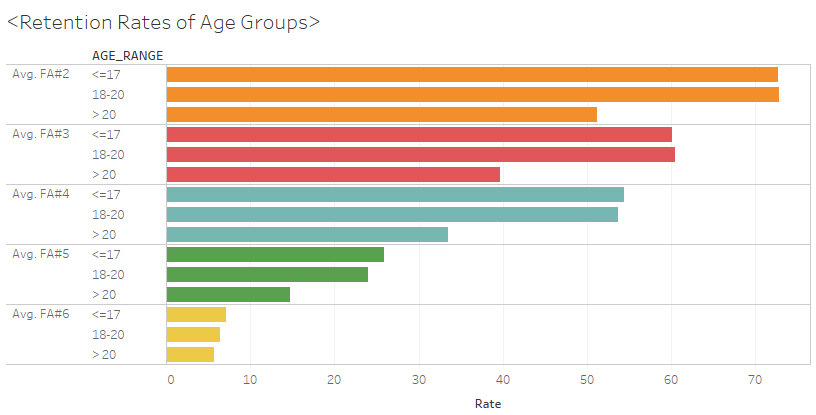}
    \caption{Retention Rates over Six Years of Age Group}
    \label{fig:retention_rates_gpa}
\end{figure}
The vast majority of FTIAC students (84\%) are between the age of 18 and 20, with just 15 percent below 17 years of age. Only one percent are older than 20 years. A closer look at the retention rates of different age groups shows that there is no difference between students below age 17 and students age 18-20, with their retention rates (73\% vs. 73\% for the second fall, 60\% vs. 60\% for the third fall, and 54\% vs.54\% for the fourth fall). Students under 20 years old achieve higher retention rates than students over 20, as seen below in Figure 5. The retention rates are 72\% vs. 51\% (second fall), 60\% vs. 39\% (third fall) and 54\% vs. 33\% (fourth fall), respectively.

\begin{tcolorbox}[colframe=black, colback=white, sharp corners=all, boxrule=1pt]
\subsubsection*{Correlation between ACT score and Retention Rates}
\textbf{Pearson correlation coefficient:} \( r = -0.866 \) \\
\textbf{P-value:} p=0.333
\end{tcolorbox}

The negative correlation coefficients of -0.866 suggest a strong inverse relationship between age groups and retention rates, indicating that younger students (below 20) tend to have higher retention rates compared to older students (over 20). However, the p-values of 0.333 indicate that these correlations are not statistically significant at the conventional alpha level of 0.05. This means that we cannot confidently conclude that there is a significant correlation between age groups and retention rates based on these data. The evidence supporting the influence of age as a factor is sufficient.

We also analyzed the variables of gender and race. Similarly to age, their correlations with student withdrawal were not statistically significant.

\subsection{Finacial Status}
The dataset contains the following key variables:\\
LOW INCOME: Indicator of whether a student is from a low-income family (Y/N).\\
FA\#2 to FA\#6: Retention rates at different follow-up fall semester.
The data is summarized in Table \ref{table:fa_scores}

\begin{table}[h!]
\centering
\begin{tabular}{lrrrrrr}
\toprule
\textbf{LOW INCOME} & \textbf{FA\#2} & \textbf{FA\#3} & \textbf{FA\#4} & \textbf{FA\#5} & \textbf{FA\#6} \\
\midrule
Y &  73.93 & 61.47 & 54.13 & 32.73 & 14.33 \\
N &  71.71 & 59.33 & 53.19 & 27.85 & 7.86 \\
\bottomrule
\end{tabular}
\caption{Retension rates by Low Income Status}
\label{table:fa_scores}
\end{table}

Contrary to some expectations and existing research, our dataset indicates that low-income students exhibit 
slightly higher retention rates at various follow-up points compared to their non-low-income peers. A plausible explanation for this trend could be targeted support programs for low-income students at Eastern Michigan University (EMU). For students who receive financial aid, they are more inclined to maintain their aid by continuing their studies.

This factor is heavily influenced by the specific policies and criteria of financial aid, which can vary significantly from one institution to another. As a result, it is difficult to generalize this finding to other schools. In the subsequent predictive models, we will exclude this factor due to its variability and the challenges in applying it consistently across different educational contexts.

\section{Predictive Model}

\subsection{Model Development}
The objective of this section is to develop predictive models for student retention, especially focusing on improving performance for students who are likely to drop out. This task is challenging due to the imbalanced nature of the dataset, where the number of retained students is significantly greater compared to the number of non-retained students. Imbalanced datasets can lead to biased models that perform well on the majority class but poorly on the minority class. To address this, we employed various techniques including advanced models like XGBoost. XGBoost is a powerful and efficient implementation of gradient boosting that often performs well on classification tasks, especially with imbalanced datasets. 

\subsubsection{Data Preparation}
The dataset includes 13995 FTIAC student records, and each record has the following feature:
\begin{itemize}
    \item \textbf{Term GPA}: Term GPA in the first fall term. 
    \item \textbf{Term Earned SCH}: Number of credit hours completed in the first fall term.
    \item \textbf{ACT Scores}: ACT composite scores.
    \item \textbf{Decision GPA}: High school GPA used for college admissions.
\end{itemize}

The target variable is the second fall retention, with a value of 1 indicating that a student is retained, and a value of 0 indicating that a student has stopped out.
Rows with missing values in the selected features and target were removed to ensure clean data. The data was split into training (80\%) and testing (20\%) sets to evaluate model performance.

\subsubsection{Model Training with XGBoost}
XGBoost is an advanced gradient boosting algorithm known for its efficiency and high performance. It is particularly effective in handling imbalanced datasets through its ability to use various parameter tuning options to handle class imbalance.
Furthermore, the SMOTE technique was applied to ensure that the minority class (retention = 1) was adequately represented.

\subsection{Results}
The model has been trained and evaluated. The performance of the model is summarized in Table \ref{table:classification_report}.

\begin{table}[h]
\centering
\begin{tabular}{|l|c|c|c|c|}
\hline
Class & Precision & Recall & F1-Score & Support \\ \hline
0 & 0.835 & 0.868 & 0.851 & 1716 \\ \hline
1 & 0.588 & 0.523 & 0.553 & 616 \\ \hline
Accuracy & & & 0.777 & 2332 \\ \hline
Macro Avg & 0.711 & 0.696 & 0.702 & 2332 \\ \hline
Weighted Avg & 0.770 & 0.777 & 0.773 & 2332 \\ \hline
\end{tabular}
\caption{Classification Report for the Trained Model}
\label{table:classification_report}
\end{table}

\subsection{Model Performance}
The overall accuracy of the model is 77.7\%, which reflects a balanced performance across both classes.
The results indicate that the model performs well in identifying the students who are retained (class 0), with high precision and recall values. For students who are not retained (class 1), the precision and recall are slightly lower, suggesting room for improvement in accurately predicting this class. \\
The macro and weighted averages provide additional insight into the model's balanced performance across different metrics.\\
The macro average provides an unweighted mean of the precision, recall, and F1-score for each class. This metric treats all classes equally, regardless of their support (the number of instances in each class). It is useful for understanding the model's performance on a per-class basis, without being influenced by class imbalance.\\
The weighted average takes into account the support of each class, providing a mean that is weighted by the number of instances in each class. This metric is particularly useful in the presence of class imbalance as it reflects the model's performance more accurately in terms of the overall population.\\
\\
The choice of model and threshold significantly impacts the recall and precision rates for students who are not retained (class 1). By selecting different thresholds, we can adjust the balance between recall and precision to suit various needs:\\

\begin{itemize}
    \item \textbf{High Precision Model}: Useful for targeted interventions where resources are limited and must be used efficiently.
\item \textbf{High Recall Model}: Suitable for broad preventive measures aiming to cover as many at-risk students as possible.\\
\end{itemize}

\subsection{Implementation}
We used data from Eastern Michigan University (EMU) to develop our predictive model, but other universities can adapt and implement this model to identify students at high risk of withdrawal. By leveraging this model, institutions can deploy early intervention strategies, offering targeted support services such as academic advising, tutoring, financial aid counseling, and engagement programs, ultimately aiming to improve student retention rates.

\section{Conclusion}
This study demonstrates the feasibility of using data-driven approaches to understand student withdrawal. By leveraging academic performance metrics, demographic information, and admission criteria, EMU can develop effective strategies to identify at-risk students in early stage and enhance overall student success.

The analysis of the characteristics and educational pathways of the stopout student population at Eastern Michigan University (EMU) has yielded valuable insights. The primary findings of this study are as follows.

\subsection{\textbf{Academic Performance as a Predictor of Retention}}  There is a strong positive correlation between academic performance metrics (such as term GPA and term earned SCH) and student retention rates. Students with higher term GPAs and earned SCH in their first fall semester are significantly more likely to stay enrolled in subsequent years. This finding underscores the critical importance of early academic success in predicting student retention.

\subsection{\textbf{Impact of Admission Criteria}}  Higher ACT scores and decision GPAs are associated with higher retention rates. This correlation indicates that students who perform well in standardized tests and high school are more likely to persist in their studies at EMU.

\subsection{\textbf{Demographic Factors}} Age, gender and race did not show statistically significant correlations with student withdrawal. These results suggest that targeted interventions may not be necessary for older students or a specific gender to help improve their retention rates.

\subsection{\textbf{Predictive Modeling}} The XGBoost model developed in this study effectively predicts the likelihood of student withdrawal based on academic performance metrics and admission criteria. The high accuracy, precision and recall values of the model indicate its robustness in identifying at-risk students.

\section{Implications for EMU and Similar Institutions}

The insights gained from this study can inform strategies to improve student retention rates at EMU and other institutions facing similar challenges. By identifying the key predictors of student withdrawal, universities can allocate resources more effectively and implement early intervention strategies to support at-risk students. 

These strategies may include the following:

\begin{itemize}
    \item Providing academic support services such as tutoring and academic advising to students with low-term GPA and earned SCH.
    \item Offering financial aid counseling to ensure that students receive adequate financial support.
    \item Designing engagement programs to help integrate students into the academic community, particularly during their first year of college.
\end{itemize}

\section{
Future Research Directions}

While this study provides a comprehensive analysis of student withdrawal at EMU, future research could explore additional factors that may influence student retention, such as finance, social integration, and external commitments (e.g., employment). Longitudinal studies could also examine the long-term impact of early interventions on student success.

In conclusion, this study highlights the importance of data-driven approaches in understanding student withdrawal. By leveraging predictive modeling and explainable AI techniques, EMU and other higher education institutions can develop effective strategies to support at-risk students and improve overall student success. The findings from this research provide a solid foundation for ongoing efforts to enhance student retention and ensure that more students achieve their academic goals.

\balance
\bibliographystyle{IEEEtran}
\bibliography{main}
\end{document}